\documentclass[letterpaper]{article} 
\usepackage{aaai2026}  
\usepackage{times}  
\usepackage{helvet}  
\usepackage{courier}  
\usepackage[hyphens]{url}  
\usepackage{graphicx} 
\usepackage{natbib}  
\usepackage{caption} 
\usepackage{algorithm}
\usepackage{algorithmic}

\usepackage{newfloat}
\usepackage{listings}
\DeclareCaptionStyle{ruled}{labelfont=normalfont,labelsep=colon,strut=off} 
\floatstyle{ruled}
\newfloat{listing}{tb}{lst}{}
\floatname{listing}{Listing}
\usepackage{amsmath,amssymb}
\usepackage{makecell}
\usepackage{multirow}
\usepackage{enumitem}
\usepackage{booktabs}
\usepackage{epsfig}
\usepackage{verbatim}
\usepackage{textcomp}

\title{How Bias Binds: Measuring Hidden Associations for Bias Control in Text-to-Image Compositions}

\title{How Bias Binds: Measuring Hidden Associations for Bias Control in Text-to-Image Compositions}

\author {
    Jeng-Lin Li\textsuperscript{\rm 1},
    Ming-Ching Chang\textsuperscript{\rm 2},
    Wei-Chao Chen\textsuperscript{\rm 1}
}
\affiliations {
    \textsuperscript{\rm 1}Inventec Corporation, No.66, Hougang St., Shilin Dist., Taipei City 111059, Taiwan\\
    \textsuperscript{\rm 2}University at Albany, State University at New York, Albany, NY, 12222, USA\\
    li.johncl@inventec.com, mchang2@albany.edu, chen.wei-chao@inventec.com
}

\usepackage{bibentry}

\begin{document}

\maketitle

\begin{abstract}
Text-to-image generative models often exhibit bias related to sensitive attributes. However, current research tends to focus narrowly on single-object prompts with limited contextual diversity. In reality, each object or attribute within a prompt can contribute to bias. For example, the prompt ``an assistant wearing a pink hat'' may reflect female-inclined biases associated with a pink hat. The neglected joint effects of the semantic binding in the prompts cause significant failures in current debiasing approaches. 
This work initiates a preliminary investigation on {\em how bias manifests under semantic binding}, where contextual associations between objects and attributes influence generative outcomes. We demonstrate that the underlying bias distribution can be amplified based on these associations. Therefore, we introduce a bias adherence score that quantifies how specific object-attribute bindings activate bias. To delve deeper, we develop a training-free context-bias control framework to explore how token decoupling can facilitate the debiasing of semantic bindings. This framework achieves over 10\% debiasing improvement in compositional generation tasks. Our analysis of bias scores across various attribute-object bindings and token decorrelation highlights a fundamental challenge: reducing bias without disrupting essential semantic relationships. These findings expose critical limitations in current debiasing approaches when applied to semantically bound contexts, underscoring the need to reassess prevailing bias mitigation strategies.
\end{abstract}


\section{Introduction}
\label{sec:intro}
Diffusion models have unlocked various applications of text-to-image (T2I) generative models. However, these models often capture spurious correlations from the training datasets, which can introduce bias during deployment in testing scenarios~\cite{wu2024stable}. Unrecognized biases embedded in models lead to skewed decision-making and societal impacts, including the reinforcement of stereotypes and concerns about fairness, which are inclined to amplify and perpetuate existing societal inequities~\cite{luccioni2023stable}.
Despite the biased correlation being observed years ago~\cite{grover2019bias}, many large-scale benchmarks still report the persistent biases in state-of-the-art T2I generative models~\cite{vice2025exploring} and biases during further distillation~\cite{luo2024faintbench}. The phenomena signify the embedded biased correlations from diverse and complex data distributions.

\begin{figure*}[t]
\centerline{
  \includegraphics[width=\linewidth]{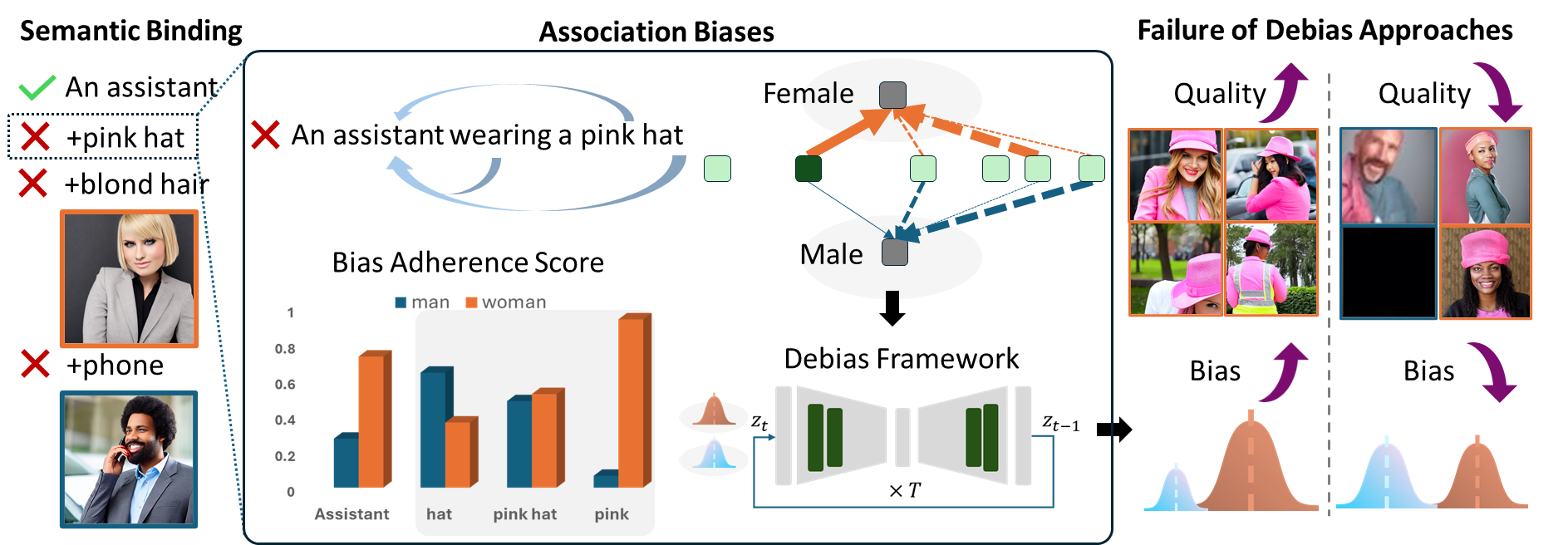} 
}
\caption{Conceptual illustration of the association bias in compositional text-to-image generation. Orange and blue boxes denote the gender bias. Current debiasing frameworks neglect the hidden bias stemming from semantic bindings, such as pink hat or blond hair, causing low-quality generation or persistent biases after debiasing.}
\label{fig:teaser}
\end{figure*}

Biases in T2I models stem from latent correlations in training data, entangled with related attributes~\cite{udandarao2024no}, making debiasing difficult without harming semantic fidelity. Recent methods address this by avoiding retraining, reducing reliance on balanced data~\cite{smith2023balancing} and counterfactual examples~\cite{jung2024counterfactually}.
Common techniques for bias mitigation include manipulating prompts~\cite{bansal2022well,ding2021cogview,chuang2023debiasing}, inserting features~\cite{li2024self}, rescaling noise guidance, and learning inclusive tokens~\cite{zhang2023iti,shrestha2024fairrag,teo2024fairqueue}. The common underlying notion of modifying the concept-bias relationship frequently compromises essential generative structures, leading to pronounced collapse. Therefore, we aim to investigate the deeper relationships between biases, objects, and attributes.

Modern bias assessment studies focus on fairness and bias metrics independent of complex prompts. 
Nevertheless, most studies focus on generating images of a single object (concept), failing to account for the increasing diversity of generation targets. For example, gender biases in occupational representations can be mitigated in simple prompts like ``a headshot of an assistant'' using current debiasing algorithms. However, in context-rich prompts such as ``a headshot of an assistant wearing a pink hat,'' these biases may persist, indicating limitations in existing debiasing techniques.
The contexts ``wearing a pink hat'' vary the token probability and amplify the spurious correlation. 
We empirically find significant failure using state-of-the-art debiasing algorithms~\cite{li2024self,teo2024fairqueue,parihar2024balancing} in the right part of Figure~\ref{fig:teaser}, which results in an unreal visual style, missing professional properties, and even violation of the Stable Diffusion Safe Checker. 
This prompts the research question: \textit{How do object and attribute bindings influence generative bias beyond the original single-object setting?}
As the first to explore this question, we focus the discussion on gender bias in occupations involving human-associated objects within compositional bias.

In this work, we quantitatively assess bias using a bias adherence score (BA-Score) in the compositional T2I generation task and reveal the insights for designing a training-free context bias control (CBC) framework. 
Our focus is on exploring how contextual biases influence the main object and impact the quality of its generation.
Figure~\ref{fig:teaser} shows the prompt ``an assistant wearing a pink hat'' producing lower-quality generations for males than females due to the underlying biases toward females. This underscores a critical consideration: the pursuit of unbiased generation may compromise reliability, as it can conflict with the inherently learnt features in the model.
Therefore, increasing importance emerges in quantifying biases in complex contexts alongside the inter-object relationship.

Our CBC framework decouples the embedding into the attribute-orthogonal embedding and the attribute-residual embedding. We use attribute-orthogonal embedding as the model input and adaptively inject the residual embedding to control the bias tendency for each generation step. During the iteration steps, we continuously measure the latent embedding distance with the attribute cluster centers. The BA-Score serves as an initialization of this bias measure, as the latent embedding is unavailable in the initial stage without performing any forward step. Our experiments demonstrate the promising bias mitigation using our CBC framework in a training-free manner and also reveal the importance of the BA-score initialization.  
We summarize our contribution as follows: 
\begin{itemize}
    \item The investigation of underexplored semantic binding biases in T2I generation, with embedding analysis uncovering key debiasing challenges.
    \item A training-free bias control framework yielding over 10\% improvements of debiasing performance without quality degradation in compositional generation.
    \item In-depth experiments to reveal various compositional scenarios that illustrate effects of underlying token correlations.
\end{itemize}


\section{Related Work}
\label{sec:related_work}

\subsection{Bias Measurements for T2I Models}
Standard evaluation protocols for T2I models have progressively integrated more complex conditions, yet still include limited bias assessments~\cite{li2024evaluating}. Given the implicit nature of bias, uncovering hidden correlations is a crucial step toward meaningful assessment.
Counterfactual reasoning has proven valuable for improving the explainability of bias evaluation~\cite{chinchure2024tibet}, while multi-aspect stereotype scoring extends traditional metrics, such as standard deviation across sensitive groups, by analyzing quantitative relationships within the latent space and across denoising steps~\cite{dehdashtian2025oasis}. Recent approaches even utilize large language models to identify open-set biases beyond predefined attributes~\cite{d2024openbias}.
Moreover, bias patterns may intersect with other persistent challenges in generative models, including hallucinations~\cite{10943752} and object omission~\cite{vice2025exploring}, further complicating reliable evaluation.

\subsection{Model Debias}
\noindent\textbf{Debiasing through model retraining}.
A foundational approach in mitigating bias in T2I models relies heavily on strategies that explicitly define targeted biases through data resampling and tailored loss functions. Mainstream research casts fairness as the distribution alignment and optimization problem to ensure fair loss update and sampling of under-represented classes~\cite{shen2024finetuning,khalafi2024constrained,zhou2024association}. 
Limited access to unbiased data has driven efforts to train unbiased models from biased datasets~\cite{kim2024training} using selective finetuning~\cite{zhao2025aim} and inpainting~\cite{hirota2024resampled} techniques.

\noindent\textbf{Debias without Retraining}.
Increasing training-free debiasing studies focus on identifying key intervention factors, including prompt enhancement, concept editing, and generating guidance. 
\textbf{Prompt enhancement} is to intuitively insert semantically fair expressions to balance the resulting attribute ratio~\cite{bansal2022well,ding2021cogview,chuang2023debiasing}. While prompt learning advances these techniques by inserting learnable inclusive tokens with either text or image references~\cite{zhang2023iti,shrestha2024fairrag}), FairQueue~\cite{teo2024fairqueue} proposes a prompt queuing mechanism to avoid the unexpected attention distortion in prompt learning. 
\textbf{Concept editing} studies~\cite{gandikota2024unified} usually identify the latent-space concept vector, termed h-vector, to manipulate the generated result. Disentangling latent features purely for sensitive attributes~\cite{shi2025dissecting} can be realized without building a classifier every time~\cite{li2024self}.
\textbf{Generating guidance} techniques comprise latent feature imputation~\cite{jung2024unified}, attention map selection~\cite{jiang2024mitigating}, and minority class sampling~\cite{kim2024rethinking}. Parihar et al. introduce the distribution guidance using an attribute distribution predictor to intervene in the latent space with the targeted distribution~\cite{parihar2024balancing}. 
These studies only considered simple prompts without compositional cases. Therefore, our work extends bias mitigation into the compositional regime, which prior works did not address.

\subsection{Compositional Text-to-Image Generation}
Compositional generation is gaining traction for tackling object omission and attribute mixing~\cite{bakr2023hrs}.
FreeCustom~\cite{ding2024freecustom} designs a multi-reference self-attention to refine the alignment between multiple provided concepts and the generated image. To reduce the inconvenience of using reference images, Hu et al. observed that text embeddings exhibit information coupling and additive properties, enabling the token merging within the same concept and disentangling multiple concepts within a prompt~\cite{hu2024token}. 
Wang et al. introduce a self-consistency guidance to refine attention maps for multi-concept attribute binding~\cite{wang2025towards}. These algorithms overlook the bias metric for evaluation, which leaves a huge risk in real-world usage. 
Recent attempts to identify object-to-gender bias have been highlighted in language models~\cite{sabir2023women} while its presence and implications in T2I generation remain overlooked.

\section{Method}
\begin{figure*}[t]
\centerline{
  \includegraphics[width=\linewidth]{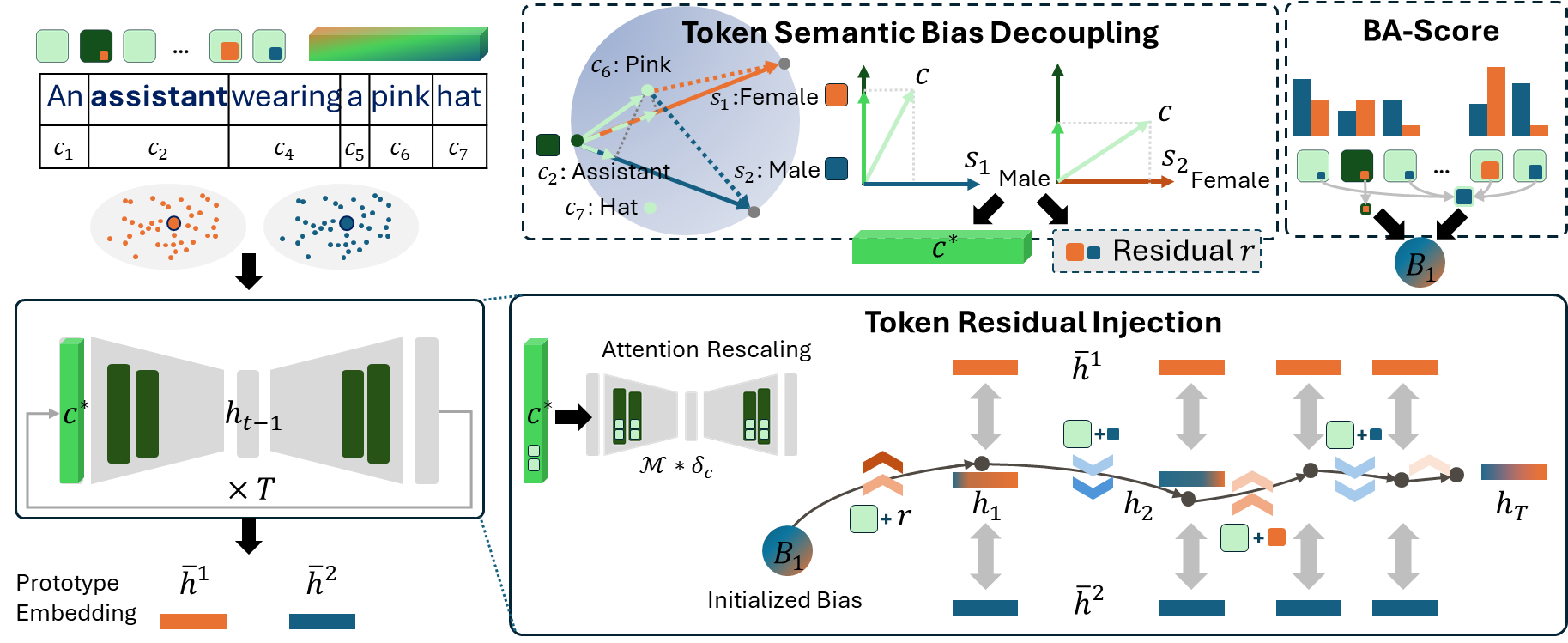} 
}
\caption{Our proposed CBC framework that decouples the token from sensitive attributes, leverages the BA-Score to measure bias tendency, and adjusts the token with sensitive attribute residuals for unbiased generation.}
\label{fig:framework}
\end{figure*}

We first explore how each concept (token) embeds with gender biases toward female or male tokens using text embedding similarity comparison. 
Figure~\ref{fig:framework} shows the overall CBC framework containing token semantic bias decoupling, BA-Score, and token residual injection. 
Our idea is to decouple the sensitive attribute-related components and control these components to balance the input embeddings at each forwarding step. The bias indicator is initialized with BA-Score and then depends on the latent embedding distance towards the gender prototypes (cluster centers of males or females). When the bias indicator skews toward a specific group, we inject attribute-related embeddings from other groups and adjust the attention to regulate the bias.
Detailed definitions and mathematical formulations are in the supplementary.

\begin{figure}[t]
\centerline{
  \includegraphics[width=\linewidth]{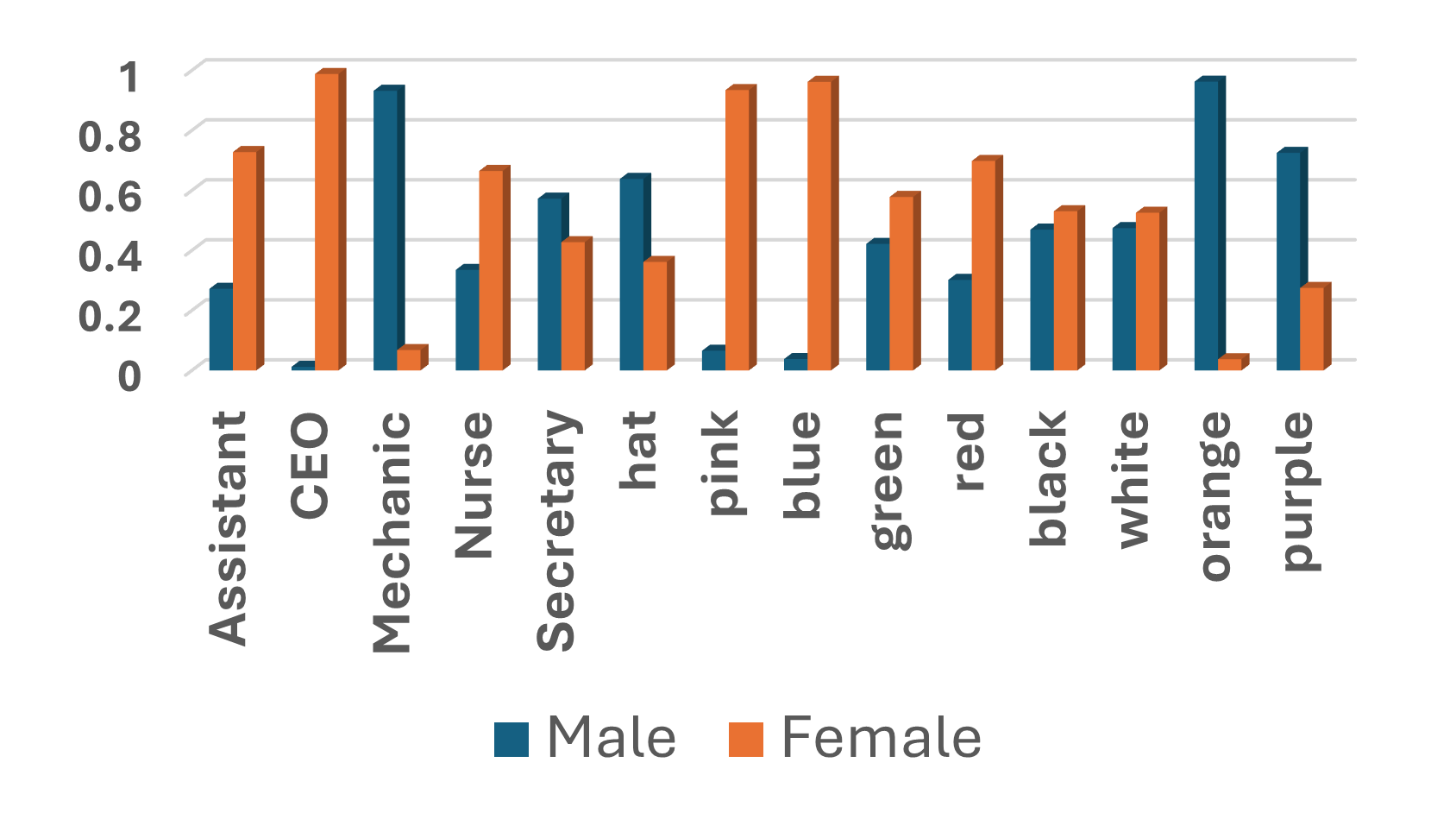} 
}
\caption{Text correlation to male or female prototype embeddings extracted via CLIP encoder.}
\label{fig:text_relation}
\end{figure}

\subsection{Motivation: Bias in Semantic Embeddings}
\label{ssec:text_relation}
We preliminarily examine the text similarity between each input token and the gender prototypes to reveal the potential bias. 
The CLIP tokenizer transforms a given prompt with $L$ tokens into text embeddings $C=\{c_1, c_2, ..., c_L\}$. Given sensitive attribute $\mathcal{S}=\{s_k\}_{k=1}^{K}$ with $K$ groups, we represent the two-group gender attributes with $\{s_1, s_2\}$.
Prototype embeddings ($\{p_1, p_2\}$) are the average embeddings over 1000 images generated by prompting ``a photo of a female'' or ``a photo of a male''. 
Considering a prompt ``an assistant'' containing $c_2=\text{``assistant''}$ that combines with ``wearing a pink hat''. The term ``pink'' ($c_6$) is endowed with high correlation to the ``woman'' prototype embedding, while ``hat'' ($c_7$) has over 0.6 correlation to the ``man'' prototype embedding in Figure~\ref{fig:text_relation}.
The combined effect of the pink hat likely shifts the bias toward more feminine features, further amplifying the inherently female-leaning latent representation of the assistant role. In contrast, the token embedding for `orange' shows a stronger correlation with `man', raising whether such a distinct bias direction could mitigate the original bias associated with 'assistant'.
The results motivate us to design a decoupling approach for effective control of sensitive components in tokens.

\subsection{Initialization: Bias Adherence Score (BA-Score)}
\label{ssec:ba_score}
Typical prompts include a unit structure, ``[main object] + [contexts]'', where contexts are composed of context tokens. Given a main object embedding $c_m$, other identified $M-1$ nouns and adjectives result in context token embeddings $C=\{c_i\}_{i\in I}$ where $I$ indicates a set of selected tokens. $I$ is regarded as a hyperparameter tuned for generation without compromising the contexts.
We define a Bias Adherence Score (BA-Score) to indicate the percentage of influence from context tokens and the main object contributing to the bias.  
We calculate the cosine similarity of context tokens to the prototype embedding $p_k$:
\begin{equation}
    B_{m,k} = \frac{
        \sum_{i=1}^{M} \mathbf{I}_{i\neq m} \exp((\cos(c_m, c_i) + \cos(p_k, c_i)) / \tau)
        }{
        \sum_{i=1}^{M} \exp((\cos(c_m, c_i) + \cos(p_k, c_i)) / \tau)
        },
\end{equation}
where $\mathbf{I}_{i\neq m}$ denotes an indicative function. The similarity from context tokens is weighted by $\cos(c_m, c_i)$, their relevance to the main object. 
BA-Score quantifies the largest deviation of sensitive attribute groups: $B_m = \max_{k} \left| \pi-B_{k,m} \right|$ where $\pi=0.5$ for targeting balance contribution. 
The more imbalanced bias contribution of the main object and contexts in an attribute group indicates the possibly strong spurious correlation from either the main object or contexts. This also implies the greater possibility of corruption after bias denoising.

\begin{table*}[t]
\centering
\setlength{\tabcolsep}{2.8pt} 
\begin{tabular}{l|ccc|ccc|ccc|ccc|ccc|ccc}
\toprule 
\multirow{2}{*}{Baseline}                 & \multicolumn{3}{c|}{Assistant} & \multicolumn{3}{c|}{CEO}  & \multicolumn{3}{c|}{Mechanic} & \multicolumn{3}{c|}{Nurse} & \multicolumn{3}{c|}{Secretary} & \multicolumn{3}{c}{Average} \\
\cline{2-19}
  & FD & VQA & AFS & FD & VQA & AFS  & FD & VQA & AFS & FD & VQA & AFS & FD & VQA & AFS & FD & VQA & AFS \\
\midrule
SD-1.5          &   0.26   &   0.58  &  \underline{0.65}  &  0.91   &  \underline{0.64}  &  0.16  &   0.96 &  \textbf{0.60} & 0.08 &   0.97  & \textbf{0.67}  & 0.06 &  0.59  & \textbf{0.73} &  0.53 &  0.69  &  \textbf{0.74} & 0.44 \\
SelfDisc  &   0.32   &  0.52  &  0.59  &   0.17  &  0.59   &  0.69  &    0.09   &  0.43   & \underline{0.58}  &      0.98   &  0.54  &  0.04 &    0.38     &   0.66  & \underline{0.64}  &  0.62  &  0.61  & 0.47 \\
DGDebias &   0.35   &  \textbf{0.61}  & 0.63  &   0.04   &  \textbf{0.66}  &  \textbf{0.78}  &    0.86   &   0.51    &   0.22  & 0.89   &  0.58  & 0.18  &   0.52      &  0.66  & 0.56  & 0.68 &  0.64  & 0.43 \\
FairQueue                       &   \underline{0.01}    &  0.32 & 0.48  &  \underline{0.03}  &  0.52  &  0.68  &    \underline{0.05}   &  0.19  &  0.32  &   0.02   &  0.12  & \underline{0.21} &   0.01   &   0.43 &   0.60  &  \textbf{0.03} &   0.33 &  \underline{0.49}  \\
CBC        &    \textbf{0.01}    &  \underline{0.60}  & \textbf{0.75} &  \textbf{0.03}   &  0.63  &  \underline{0.76} &    \textbf{0.02}  &   \underline{0.59}     &  \textbf{0.74}   &    \textbf{0.01}   & \underline{0.66}  &  \textbf{0.79} & \textbf{0.01}  &  \underline{0.70} &  \textbf{0.82} &    \underline{0.04}  &  \underline{0.68}  &  \textbf{0.80} \\
\midrule\midrule
\multirow{2}{*}{Composition}                 & \multicolumn{3}{c|}{Assistant} & \multicolumn{3}{c|}{CEO}  & \multicolumn{3}{c|}{Mechanic} & \multicolumn{3}{c|}{Nurse} & \multicolumn{3}{c|}{Secretary} & \multicolumn{3}{c}{Average} \\
\cline{2-19}
  & FD & VQA & AFS & FD & VQA & AFS  & FD & VQA & AFS & FD & VQA & AFS & FD & VQA & AFS & FD & VQA & AFS \\
\midrule
SD-1.5    & 0.35 & \underline{0.60} & 0.62 & 0.92 & \underline{0.64} & 0.14 & 0.69 & \textbf{0.62} & 0.35 & 0.63 & \textbf{0.67} & 0.40 & 0.68 & \underline{0.65} & 0.42 & 0.69 & \textbf{0.63} & 0.41 \\
SelfDisc  & 0.37 & 0.54 & 0.58 & 0.37 & 0.55 & 0.59 & 0.42 & 0.56 & 0.56 & 0.94 & 0.60 & 0.14 & 0.76 & 0.63 & 0.37 & 0.61 & 0.58 & 0.42 \\
DGDebias  & 0.55 & 0.43 & 0.43 & 0.61 & 0.55 & 0.43 & 0.93 & 0.55 & 0.12 & 0.90 & 0.51 & 0.17 & 0.78 & 0.62 & 0.31 & 0.73 & 0.62 & 0.35 \\
FairQueue & \underline{0.03} & 0.52 & \underline{0.67} & \underline{0.04} & 0.58 & \underline{0.72} & \underline{0.11} & 0.44 & \underline{0.59} & \underline{0.04} & 0.40 & \underline{0.54} & \underline{0.04} & 0.38 & \underline{0.53} & \underline{0.04} & 0.45 & \underline{0.60} \\
CBC       & \textbf{0.03} & \textbf{0.63} & \textbf{0.76} & \textbf{0.03} & \textbf{0.65} & \textbf{0.78} & \textbf{0.04} & \textbf{0.62} & \textbf{0.75} & \textbf{0.03} & \underline{0.60} & \textbf{0.77} & \textbf{0.03} & \textbf{0.66} & \textbf{0.79} & \textbf{0.04} & \underline{0.62} & \textbf{0.75} \\
\bottomrule
\end{tabular}
\caption{FD ($\downarrow$), VQA ($\uparrow$), AFS ($\uparrow$) results of SD-1.5 and debiasing approaches. Full results are in the supplementary.}
\label{table:main_results}
\end{table*}

\subsection{Context Bias Control (CBC)}
\label{ssec:cbc}
We decouple the context token’s embedding and remove the sensitive attribute components, yielding a \textit{attribute-decoupled} token embedding and the corresponding attribute residual vector. 
During diffusion, we continually measure the bias using the latent vectors and inject the residual vector of other attributes into the prompt token embedding to balance the generation for the next time step.

\noindent\textbf{Token Semantic Bias Decoupling}
We decouple the context tokens from sensitive attributes by orthogonizing text prompt embeddings, and derive attribute-orthogonal token embeddings $C^*$. 
The Schmidt orthogonalization projects a token embedding $c$ to a new direction orthogonal to the sensitive attribute embedding $s_k$: 
\begin{equation}
    c^* = c-r_k = c - \frac{\langle c, s_k \rangle}{\langle s_k, s_k \rangle}s_k,
\label{eq:orthogonal}
\end{equation}
where $\langle c, s_k \rangle$ denotes inner product between $c$ and $s_k$. We term $r_k$ as the attribute-specific residual vector that entails attribute information adhering to $c$. The projection ensures that the new concept embedding $c^*$ eliminates dependent information in $s$. Therefore, the following T2I generation regards pure semantics from the concept rather than spurious correlation from sensitive attributes.

In contrast to prior works that directly manipulate latent vectors or introduce loss-based guidance to steer the main concept semantics toward a less sensitive space~\cite{gandikota2024unified,shi2025dissecting}, we propose decoupling the influences of individual objects and attributes before intervening in the T2I generation process. The model thus forms a pure relation in attention maps from the attribute-orthogonal input embeddings rather than conceptually mixed text embeddings. The decoupled residual vector for the sensitive attributes is preserved for the dynamic denoising steps.

\noindent\textbf{Token Residual Injection}
We inject the residual embedding of the context tokens to adjust the bias tendency. The injection depends on the measure of current bias tendency via a bias deviation score. We inject the average residual embedding from the other attributes if we find the current generating step deviates from an attribute group $s_k$.  
Considering the residual embedding of other sensitive attributes written as $\overline{r}=\frac{1}{K-1}\sum_{j\neq k}{r_j}$, we can use a weighting factor $\delta_r$ to add control over the conditional text embedding $c^*_{t} = \delta_r*\overline{r} + (1-\delta_r)*c^*_{t-1}$ for the next generative iteration.
For the bias deviation score, we adapt the calculation of the BA-score to the latent space from the original semantic space. That is, measuring the BA-score with latent embedding $h_{t-1}$ and the latent prototype embedding $\overline{h}^{k}$ for sensitive attribute $k$. It is noteworthy that $\overline{h}^{k}$ is extracted by a contrastive network module trained on 1000 images from each attribute group~\cite{li2024self}. To generalize the $\overline{h}^{k}$ for all the time steps, we include the randomly sampled images from each time step during the contrastive training. 
We utilize text embedding based BA-Score for initialization and latent-space BA-Score for adjusting the residual direction in the intermediate time steps. This residual not only control the bias but componsate for the corrupted spurious correlation due to token decoupling.

Token residual injection may confront numerical issues and unstable convergence in that the injection also influences other compositional attributes and objects in the cross-attention calculation. 
Therefore, we introduce an attention rescaling mechanism to adjust the attention weight of the injected tokens.  
Given the $i^{th}$ token is injected with the residual embedding, its attention vector $\mathcal{M}_i\in \mathbb{R}^{1\times L}$ for $L$ input tokens is rescaled by a weight scalar $\delta_c$. The calibrated attention mask $\mathcal{M}^*_i$ thus emphasizes more where the token relates more to the subject of the prompt. 
$\mathcal{M}^*_i=w(t)\delta_c\mathcal{M}_i$
where $w(t)=1-\frac{t}{T}$ is time-aware strength attenuated function.

\section{Experiments}
\label{sec:experiment}

\noindent
{\bf Dataset:}
We evaluate the occupation bias on Winobias~\cite{zhao2018winobias} benchmark that includes 36 professions known to exhibit gender biases. 
Each prompt template, ``a head of a [occupation] [semantic binding]'', is used to generate 200 images for each occupation evaluation. Full composition settings are listed in the supplementary.

\noindent
{\bf Evaluation metric:}
We consider the metrics: 1) Faireness Discrepancy (FD) used to assess the bias regarding a sensitive attribute~\cite{teo2024fairqueue,li2024self,orgad2023editing}, 2) VQAScore~\cite{lin2024evaluating} (VQA) used for text alignment of the image generation, 3) Alignment-aware Fairness Score (AFS) to reflect the intrinsic debiasing results without comprising text-alignment quality. 
We define the score, $\text{AFS}=\frac{2*(1-\text{FD})*\text{VQA}}{(1-\text{FD}) + \text{VQA}}$. The highest score of $\text{AFS}=1$ can be achieved only if $\text{FD}=0$ and $\text{VQA}=1$. Higher harmonic mean in this metric suggests a better balancing of the bias-utility tradeoff~\cite{zhao2022inherent}.
VQAScore has shown superior alignment to human evaluation for image generation faithfulness~\cite{li2024evaluating}.
The implementation of FD depends on CLIP ViT-L-14, and the VQAScore is calculated with the outputs from LLaVA-1.5~\cite{liu2024improved}.

\noindent
{\bf Compared Models and Configurations:}
We use pretrained Stable Diffusion v1.5 as our base model to generate images with 512$\times$512 resolution using a single Nvidia A40 GPU. The generation process is implemented with a guidance scale equal to 7.5 in 50 steps. For attention weighting control, $\delta_c=2$ and $\delta_r=0.2$ are empirically optimal.  
The compared debiasing approaches mainstream debiasing concepts include concept editing (SelfDisc~\cite{li2024self}), generative guidance (DGDebias~\cite{parihar2024balancing}), and prompt enhancement (FairQueue~\cite{teo2024fairqueue}).

\subsection{Association Bias Mitigation Results}

\noindent\textbf{Quantitative comparison}.
Semantic bindings using the prompt ``wearing a [object]'' and ``wearing a [color] [object]''. The verb is adapted based on the object. For example, carrying a briefcase is preferred over wearing one.  
Table~\ref{table:main_results} demonstrates the average results for a set of professions observed to be biased in the SD model, and that are added with semantic bindings. 
The results reveal hidden restrictions of SoTA debiasing approaches, as SelfDisc and DGDebias, which are ineffective and obtain high FD scores in compositional generation scenarios. The average compositional results present 0.41, 0.42, and 0.35 AFS scores. However, FairQueue manifests a typical tradeoff with low FD scores (0.04) and VQAScore (0.45), and its AFS (0.60) still performs worse than our CBC framework.
The debiasing forces gender concepts to balance, betraying the source training distribution, and thus compromising the generated image quality. Our proposed CBC framework achieves 0.04 FD, 0.62 VQA, and 0.75 AFS in compositional prompts. The results are coherent with our hypothesis that compositional bias is associated with semantic correlation. The context components are vulnerable to compulsory bias removal and significantly degrade visual quality. The design of continuous token residual injection alleviates the problem by delicately adjusting the contexts in balancing the bias tendency and context dependency.

\begin{figure*}[t]
\centerline{
  \includegraphics[width=\linewidth]{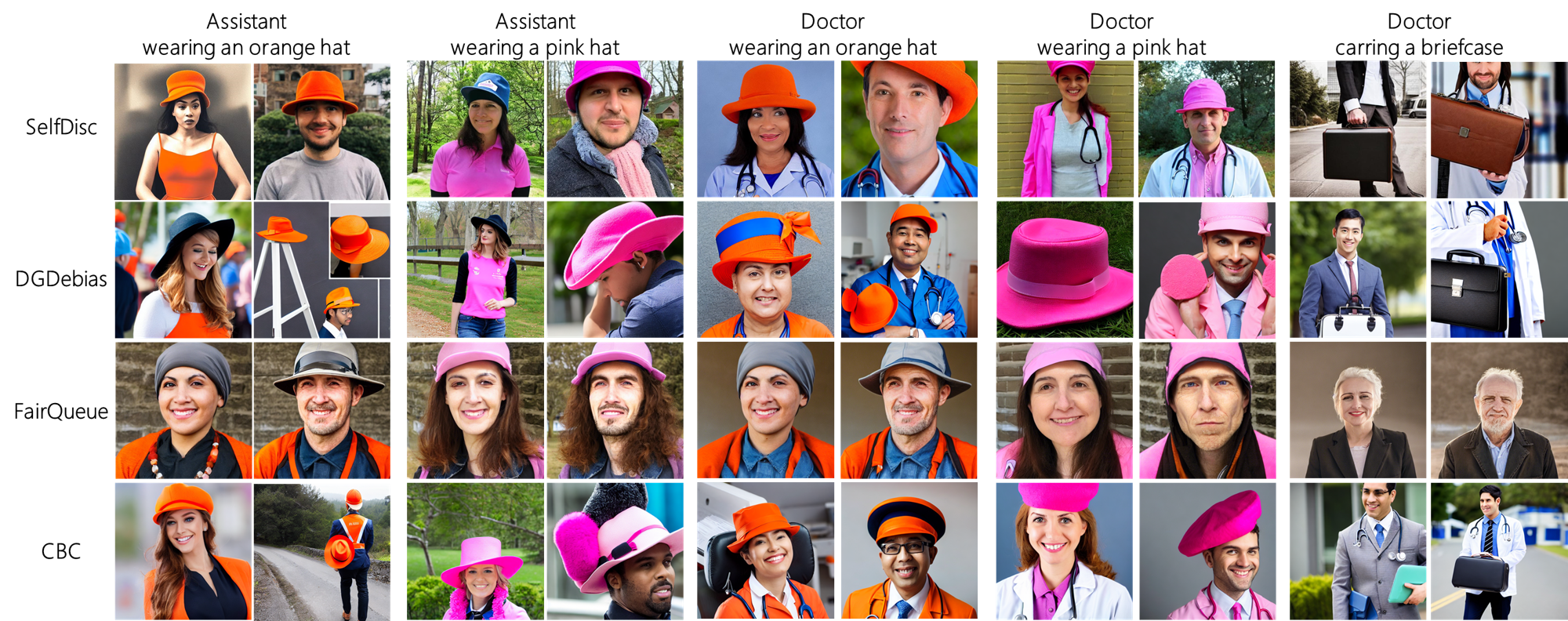} 
}
\caption{Generated images using different debiasing approaches.}
\label{fig:case_study}
\end{figure*}

\begin{figure*}[t]
\centerline{
  \includegraphics[width=\linewidth]{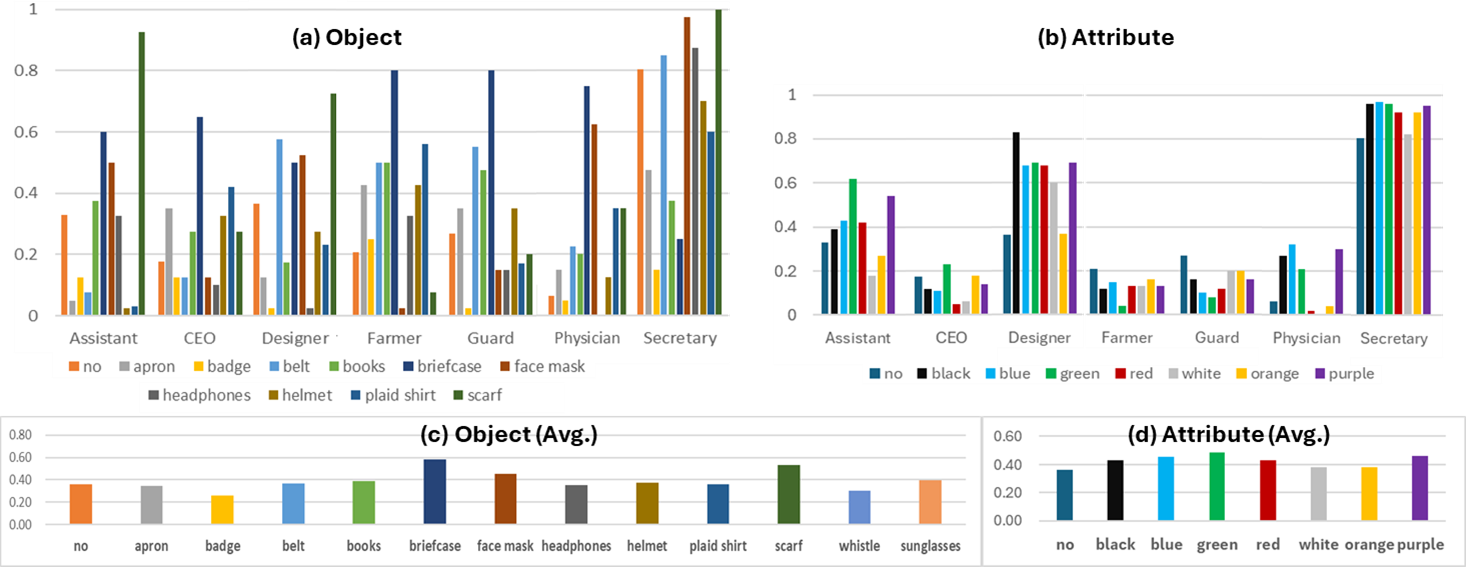} 
}
\caption{Bias results (FD) in binding conditions with (a) objects and (b) color attributes. The corresponding averaged results for objects and attributes are presented in (c) and (d). Results from other professions are reported in the supplementary. }
\label{fig:complex_composition}
\end{figure*}

\begin{figure*}[t]
\centerline{
  \includegraphics[width=\linewidth]{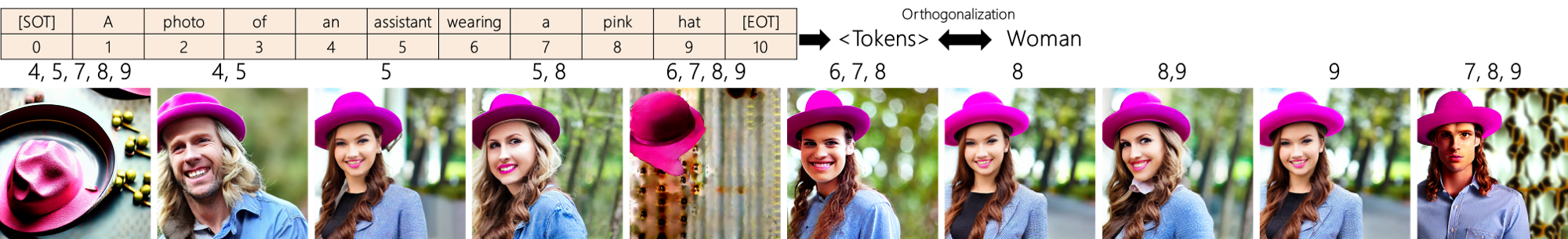} 
}
\caption{Visualization of the decoupling effects on different tokens. The annotated number of tokens is decoupled to be orthogonal to the ``woman'' token. }
\label{fig:decorrelated_effects}
\end{figure*}

\noindent\textbf{Qualitative comparison}.
The generated images in Figure~\ref{fig:case_study} show the superior text alignment quality from our proposed framework compared to other debiasing approaches. Looking into the images generated from DGDebias, the second image (orange hat assistant) shows a typical failure case that introduces redundant hats and irregular components. The DGDebias model erroneously assigns a dark hat and orange clothing to the woman in the first image, highlighting an issue of attribute mixing. FairQueue exhibits a different limitation where many of its generated images lack recognizable features associated with the intended professions. This issue is particularly evident in prompts for doctors, where the resulting images closely resemble those of assistants. Other debiasing methods typically rely on the presence of a stethoscope to fulfill the ``doctor'' prompt.  
These results suggest that current debiasing approaches struggle with properly performing composition generation. FairQueue often neglects the prompt's intended subject, while DGDebias overcorrects borderline cases, leading to visible image distortion.
SelfDisc demonstrates relatively stronger text-to-image alignment, although its FD scores remain suboptimal, as shown in Table~\ref{table:main_results}.

\subsection{Ablation and Compositional Effect Analysis}
\noindent\textbf{Effects of BA-Score and hyper-parameters}.
Table~\ref{tab:ablation} presents the ablation results for our CBC framework. 
Discarding BA-Score initialization leads to 7\% AFS drops and using simple semantic similarity for initialization brings 11\% AFS degradation. We emprically derive the best hyper-parameters ($\delta_c=2$ and $\delta_r=0.2$) which shows superior results over other values. The ablation study indicates the effectiveness of initialization with BA-Score that highly impacts the control of generating steps.

\noindent\textbf{Decoupling Effects}.
We examine the effects of token decoupling in terms of the underlying bias. Figure~\ref{fig:decorrelated_effects} illustrates the generated images with different selected tokens being decorrelated. Given the prompt ``a photo of an assistant wearing a pink hat'' with a high BA-Score due to the ``assistant'' and ``pink hat'', we project the selected tokens to be orthogonal to the ``woman'' token embedding. 
As a result, decoupling ``an assistant'' (tokens 4,5) and ``a pink hat'' (tokens 7,8,9) yields male characteristics, while several other combinations fail to change the bias. Performing orthogonization on ``assistant'', ``pink'', or ``hat'' cannot significantly modify the bias tendency. This phenomenon echoes an observed effect in composition T2I research called token information leakage. The information of the noun might leak to proximate tokens due to cross-attention calculation. Intriguingly, overdecorrelation degrades the semantics. For example, simultaneously decoupling five tokens for "an assistant" and "a pink hat" (4, 5, 7, 8, 9) results in an image without a human but with a hat. A similar result occurs when decorrelating ``wearing a pink hat'', which misleads the model to ignore the wearing action towards a human. This is because the semantic space of ``woman'' is always associated with humans. The decoupling removes the semantic relations to the woman, bringing about a side effect of lower confidence in generating humans. 
Therefore, our empirical finding suggests that keeping the main subject, such as an assistant, and debiasing the associated effects on the compositional attributes and objects can alleviate the semantic decoupling.

\begin{table}[t]
\centering
\begin{tabular}{l|ccc}
\toprule
                          & FD   & VQA  & AFS  \\
\midrule            
CBC ($\delta_c$=2, $\delta_r$=0.2)  & 0.04 & 0.62 & 0.75 \\
without BA-Score Initialization     & 0.16 & 0.58 & 0.68 \\
Semantic-Score Initialization   & 0.19 & 0.53 & 0.64 \\
$\delta_c$=1              & 0.08 & 0.61 & 0.73 \\
$\delta_c$=5              & 0.11 & 0.59 & 0.71 \\
$\delta_r$=0.3            & 0.04 & 0.60 & 0.74 \\
$\delta_r$=0.5            & 0.12 & 0.58 & 0.70 \\
\bottomrule
\end{tabular}
\caption{Ablation results for BA initialization, $\delta_c$ and $\delta_r$.}
\label{tab:ablation}
\end{table}

\noindent\textbf{Object Binding Prompts}. We regard binding objects and use the prompt ``a photo of a [profession] wearing a [object]'', where the objects are provided by prompting GPT-3.5 for common human accessories. The verb is adapted based on the object. For example, carrying a briefcase is preferred over wearing one. In Figure~\ref{fig:complex_composition} (b), adding “carrying a briefcase” reduces bias in female-leaning roles like secretary. In contrast, “wearing a scarf” sharply increases bias, e.g., the assistant’s bias ratio jumps from below 0.6 to over 0.9.

\noindent\textbf{Attribute Binding Prompts}. We include colors: blue, red, green, orange, black, white, pink, in the prompt ``a photo of a [profession] wearing a [color] hat''.
Notably, changing hat color alone can shift bias across professions. For example, designers show amplified bias with black hats, while physicians exhibit higher FD with blue hats, likely reflecting the common use of surgical caps. These trends align with social stereotypes and dataset distributions.
Intriguingly, the green hat triggers substantial bias increases among multiple white-collar professions, e.g., assistants, typically not associated with this color. Conversely, professions like farmers and guards exhibit less bias when paired with green hats, suggesting less distortion from established associations. Color specifications can intensify latent biases, amplifying or attenuating them depending on the subject’s occupational context. These results are compared with the pre-inference text embedding relationship in the supplementary.

\noindent\textbf{Associations to Other Spurious Correlation}.
We observe that the successfully generated images shown in Figure~\ref{fig:case_study} potentially imply other spurious correlations. For instance, the prompt, ``An assistant wearing a pink hat'' frequently produces outdoor scenes with green fields and trees, likely reflecting the rarity of pink hats in office settings. 
A similar trend is observed in SelfDisc’s output for a doctor wearing a pink hat, highlighting persistent spurious associations across different professions and debiasing strategies.
Moreover, the failed cases for ``wearing a briefcase'' in Figure~\ref{fig:case_study} are accompanied by a suit, regardless of the debiasing results. 
When evaluating prompt alignment, the token correlation can suggest either a meaningful association or an unintended bias, depending on the sensitivity of the attribute involved. These observations highlight a research direction: stratify less sensitive attributes and leverage their underlying correlations to support realism, thereby alleviating the diffusion model's struggles with underrepresented bindings.

\section{Conclusion}
In this work, we pioneer the estimation of bias adherence in compositional T2I generation, where existing debiasing methods often fall short. By measuring bias scores across bound objects and attributes, our context-bias control framework decouples token embeddings and dynamically adjusts residuals to steer generation toward an unbiased and less corrupted direction. Experimental results demonstrate that our approach effectively mitigates bias while maintaining strong text-image alignment.
Further experiments reveal the associated biases from different attributes and objects, which opens up an unexplored avenue for broader debiasing scenarios. Our findings on complex compositions and decorrelation results suggest the potential to leverage token relations for debiasing. Future works include constructing large-scale benchmarks for debiasing compositional T2I tasks and investigating other compositional approaches for sophisticated attention guidance. 

\bibliography{aaai2026}

\end{document}